# Robust Fusion of LiDAR and Wide-Angle Camera Data for Autonomous Mobile Robots

**Varuna De Silva\*, Jamie Roche and Ahmet Kondoz**

Institute for Digital Technologies, Loughborough University, London E15 2GZ, UK
\* Correspondence: V.D.De-Silva@lboro.ac.uk; Tel.: +44-742-683-xxxx

**Abstract—**Autonomous robots that assist humans in day to day living tasks are becoming increasingly popular. Autonomous mobile robots operate by sensing and perceiving their surrounding environment to make accurate driving decisions. A combination of several different sensors such as LiDAR, radar, ultrasound sensors and cameras are utilized to sense the surrounding environment of autonomous vehicles. These heterogeneous sensors simultaneously capture various physical attributes of the environment. Such multimodality and redundancy of sensing need to be positively utilized for reliable and consistent perception of the environment through sensor data fusion. However, these multimodal sensor data streams are different from each other in many ways, such as temporal and spatial resolution, data format, and geometric alignment. For the subsequent perception algorithms to utilize the diversity offered by multimodal sensing, the data streams need to be spatially, geometrically and temporally aligned with each other. In this paper, we address the problem of fusing the outputs of a Light Detection and Ranging (LiDAR) scanner and a wide-angle monocular image sensor for free space detection. The outputs of LiDAR scanner and the image sensor are of different spatial resolutions and need to be aligned with each other. A geometrical model is used to spatially align the two sensor outputs, followed by a Gaussian Process (GP) regression-based resolution matching algorithm to interpolate the missing data with quantifiable uncertainty. The results indicate that the proposed sensor data fusion framework significantly aids the subsequent perception steps, as illustrated by the performance improvement of a uncertainty aware free space detection algorithm.

**Keywords:** sensor data fusion; depth sensing; LiDAR; Gaussian Process regression; free space detection; autonomous vehicles; assistive robots

## 1. Introduction

Assistive autonomous robots that help humans in day-to-day tasks are becoming increasingly popular in domestic and industrial applications. Indoor cleaning robots [1,2], surveillance robots [3], lawn mowing and maintenance robots [4,5] and indoor personal assistant vehicles for the disabled [6] are but a few applications of autonomous assistive robots on the horizon. In the near future, one of the most popular consumer applications of mobile robots will be in the form of self-driving passenger/cargo vehicles [7]. While several major automobile manufacturers have set targets to launch commercially available fully autonomous driverless vehicles by 2020, vehicles that are adequately capable to roam without the need for human intervention, are still a distant reality that requires extensive research effort for realization [8].

A typical autonomous mobile robot/vehicle is composed of three major technological components: a mapping system that is responsible for sensing and understanding the objects in the surrounding environment; a localization system with which the robot comes to know its current location at any given time, and the third component responsible for the driving policy. The driving policy refers to the decision-making capability of the autonomous robot when faced with various situations, such as negotiating with human agents and other robots. Effective environment mapping
........

is crucial accurate localization and driving decision making of the mobile robot. Throughout the paper, we will use the term autonomous mobile robots and autonomous vehicles interchangeably.

Current prototypes of autonomous mobile robots (also widely referred in the literature as autonomous vehicles) [9] utilise multiple different sensors, such as Light Imaging Detection and Ranging (LiDAR), radars, imaging and ultrasound sensors to map and understand their surroundings. Radar is used for long-range sensing, while ultrasound sensors are effective at very short ranges. Imaging sensors are often used to detect objects, traffic signals, lane markings and surrounding pedestrians and vehicles. Often, these prototypes rely on LiDAR sensors or stereo cameras to map the surrounding environment in 3-dimensions (3D). Data generated from each sensor need to be interpreted accurately for satisfactory operation of autonomous vehicles. The precision of operation of an autonomous vehicle is, thus, limited by the reliability of the associated sensors. Each type of sensor has its own limitations, for example, LiDAR sensor readings are often affected by weather phenomena such as rain, fog or snow [10]. Furthermore, the resolution of a typical LiDAR sensor is quite limited as compared to RGB-cameras. In comparison, stereo camera-based dense depth estimation is limited by its baseline distance [11]. Therefore, for accurate operation an autonomous mobile robot typically relies on more than one type of sensor.

The diversity offered by multiple sensors can positively contribute to the perception of the sensed data. The effective alignment (either spatially, geometrically or temporally) of multiple heterogeneous sensor streams, and utilization of the diversity offered by multimodal sensing is referred to as sensor data fusion [12]. Sensor data fusion is not only relevant to autonomous vehicles [13], but also applicable in different applications such as surveillance [14], smart guiding glasses [15] and hand gesture recognition [16]. Overcoming heterogeneity of different sensors through effective utilization of redundancy across the sensors is the key to fusing different sensor streams.

Wide-angle cameras are increasingly becoming popular in different applications. Unlike standard cameras, wide angle cameras provide the capability to capture a broad area of the world with as few sensors as possible. This is advantageous from a cost perspective as well as from a system complexity perspective. If wide-angle cameras can be utilized effectively in mobile robots it will pave way for more compact and cost-effective robots. In this paper, we investigate indoor mobile robot navigation by fusing distance data gathered by a LiDAR sensor, with the luminance data from a wide-angle imaging sensor. The data from the LiDAR comes in the form of a 3D point cloud, whereas wide-angle camera captures the scene from a larger visual angle (typically >180°). Recently, LiDAR data and wide-angle visual data were fused for odometry and mapping of indoor environments [17]. Most work that involves camera and LiDAR fusion often focuses on extrinsic calibration of the two sensors to align the data [18,19]. However, data fusion goes beyond extrinsic calibration and involves resolution matching, handling missing data and accounting for variable uncertainties in different data sources. The objective of this paper is to address the issues of resolution mismatch and uncertainty in data sources while fusing wide angle camera and LiDAR data.

LiDAR and stereo camera fusion is often tackled through utilization of the common dimension of depth in the two modalities. In comparison, fusing LiDAR with wide-angle luminance data is non-trivial as there is no common dimension of depth, as there is no way to capture depth in a wide-angle camera. To overcome above challenges, in this paper we try to address the problem of fusing LiDAR data with wide-angle camera. Furthermore, the technique goes beyond a simple geometric calibration by developing a robust fusion algorithm, which enable the robot to make decisions under uncertainty. We illustrate the effectiveness of our approach with a free space detection algorithm, which utilizes the fused data to understand areas in the world that the robot can navigate to without colliding with any obstacle.

The rest of this paper is organised as follows: Section 2 provides an overview of the related work and associated challenges. The framework for fusion of LiDAR and Imaging sensor data are presented in Section 3. Section 4 describes the experimental framework and discussion of the free space detection results. Finally, we conclude the paper in Section 5, with some references to possible future work.

## 2. Sensor Data Fusion for Computer Vision

A review of the literature relevant to the contributions of this paper is presented in this section, followed by the positioning of the current contribution. This section is organized in three sections: the need for data fusion and challenges it poses, relevant work in LiDAR and camera data fusion and finally on challenges addressed by the current work within the scope of driverless vehicles.

### 2.1. Challenges in Multimodal Data Fusion

Information about a system can be obtained from different types of instruments, measurement techniques and sensors. Sensing a system using heterogeneous acquisition mechanisms is referred to as multimodal sensing [20]. Multimodal sensing is necessary, because a single modality cannot usually capture complete knowledge of a rich natural phenomena. Data fusion is the process by which multimodal data streams are jointly analysed to capture knowledge of a certain system.

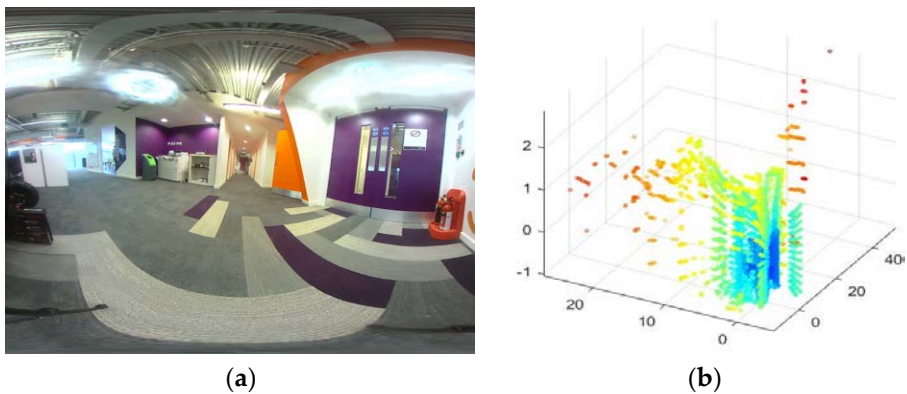

(**a**)                                                                  (**b**)

**Figure 1.** Examples of sensor data to be fused. (**a**) Image from the wide angle camera; (**b**) 3D point cloud from LiDAR.

Lahat et al. [20], identified several challenges that are imposed by multimodal data. These challenges can be broadly categorized into two segments: challenges at the acquisition level and challenges due to uncertainty in the data sources. Challenges due to data acquisition level problems include: differences in physical units of measurement (non-commensurability), differences in sampling resolutions, and differences in spatio-temporal alignment. The uncertainty in data sources also pose challenges that include: noise such as calibration errors, quantization errors or precision losses, differences in reliability of data sources, inconsistent data and missing values.

The above challenges discussed by Lahat et al. [20], were identified by considering a multitude of applications. In the next subsection, we will discuss specific challenges associated with fusing LiDAR and imaging data.

### 2.2. Fusion of LiDAR and Different Types of Imaging Data

LiDAR data can be fused with different types of imaging sensor data to cater for a range of applications. An example of two types of data, i.e. from wide angle camera and the LiDAR are illustrated in Figure 1. Terrain mapping is a popular application of LiDAR data that uses an aerial borne LiDAR scanner to identify various ground objects such as buildings or vehicles. The independent use of LiDAR scanner proves challenging in such applications due to obstructions and occlusions caused by vegetation. Therefore, while LiDAR exhibits good height measurement accuracy, it lacks in horizontal segmentation capability to delineate the building boundaries. A graph based data driven method of fusing LiDAR data and multi-spectral imagery was proposed in. Authors in [21] propose, a connected component analysis and clustering of the components to come up with a more accurate segmentation algorithm.

In a substantial body of literature, LiDAR and image data fusion is considered as an extrinsic calibration process. Here fusion is regarded as the process of rigid body transformation between the two sensors' coordinate systems [18]. For the purpose of extrinsic calibration, an external object, such as a trihedral calibration rig [22,23], a circle [24], a board pattern [25,26] or a checkerboard pattern [27–29], is used as a target to match the correspondences between the two sensors. Li et al. proposes a calibration technique to estimate the transformation matrix that can then be used to fuse a motorized 2D laser scanner with a monocular image [19]. While such methods, yield accurate alignment, they do not address the issues related to uncertainty of sensor readings.

The problem of LiDAR and imaging data fusion can be approached as a camera pose estimation problem, where the relationship between 3D LIDAR coordinates and 2D image coordinates is characterised by camera parameters such as position, orientation, and focal length. In [30], the authors propose an information-theoretic similarity measure to automatically register 2D-Optical imagery with 3D LiDAR scans by searching for a suitable camera transformation matrix. LiDAR and optical image fusion is used in [30] for creating 3D virtual reality models of urban scenes.

The fusion of 3D-LiDAR data with stereoscopic images is addressed in [31]. The advantage of stereoscopic depth estimation is its capability to produce dense depth maps of the surroundings by utilising stereo matching techniques. However, the dense stereo depth estimation is computationally quite complex. This is due to the requirement of matching corresponding points in the stereo images. Furthermore, dense depth estimation using stereo images suffer from the limited dynamic range of the image sensors, for instance, due to the saturation of pixel values in bright regions [32].

Another, drawback of stereo based depth estimation is the limited range of depth sensing. LiDAR scanning on the other hand provides a utility to measure depth at high accuracies, albeit at lower point resolutions compared with imaging sensors. The authors of [31] proposed a probabilistic framework to fuse sparse LiDAR data with stereoscopic images, which is aimed at real-time 3D perception of environments for mobile robots and autonomous vehicles. An important attribute of probabilistic methods, such as in [31] is that it represents the uncertainty of estimated depth values.

### 2.3. Challenges in Data Fusion Addressed in this Paper

In this paper, we consider LiDAR and imaging sensor data fusion in the context of autonomous vehicles. Autonomous vehicles as an application pose significant challenges for data driven decision making due to the associated safety requirements. For reliable operation, decisions in autonomous vehicles need to be made by considering all the multimodal sensor data they acquire. Furthermore, the decisions must be made in the light of the uncertainties associated with both data acquisition methods, and the utilized pre-processing algorithms.

This paper addresses two fundamental issues surrounding sensor data fusion, namely the resolution difference in heterogeneous sensors and making sense of heterogeneous sensor data streams while accounting for varying uncertainties in the data sources. Apart from being different from previous contributions in the type of sensors used for data fusion, our motivations for this paper are two-fold: Firstly, we are interested in developing a more robust approach for data fusion, which accounts for uncertainty in the fusion algorithm. This will enable the subsequent perception tasks in an autonomous vehicle to operate more reliably. Secondly, we envisage situations in the future, where autonomous vehicles will be exchanging useful sensor data between each other. In such situations it would be impractical for extrinsic calibration methods to work, because there are inevitable per-unit variations that exist between sensors due to manufacturing variations. Based on the above premises, we propose a robust framework for data fusion with minimal calibration.

## 3. The Proposed Algorithm for LiDAR and Wide-Angle Camera Fusion

To address the challenges presented above, in this section we propose a framework for data fusion. This section describes the proposed algorithm for fusion of LiDAR data with a wide-angle imaging sensor. The organization of the section is as follows: in Section 3.1 the geometric model for alignment of the two sensor types are presented, followed by Gaussian process-based matching of resolutions of the two sensors, in Section 3.2.

### 3.1. Geometric Alignment of LiDAR and Camera Data

The first step of the data fusion algorithm is to geometrically align the data points of the LiDAR output and the 360° camera. The purpose of the geometric alignment is to find the corresponding pixel in the camera output for each data point output by the LiDAR sensor. For the purpose of this derivation, consider an object $O$ of height $Ho$ at distance $D$ from the robot. The sensor setup is graphically illustrated in Figure 2 and the horizontal alignment of the sensors are depicted in Figure 3.

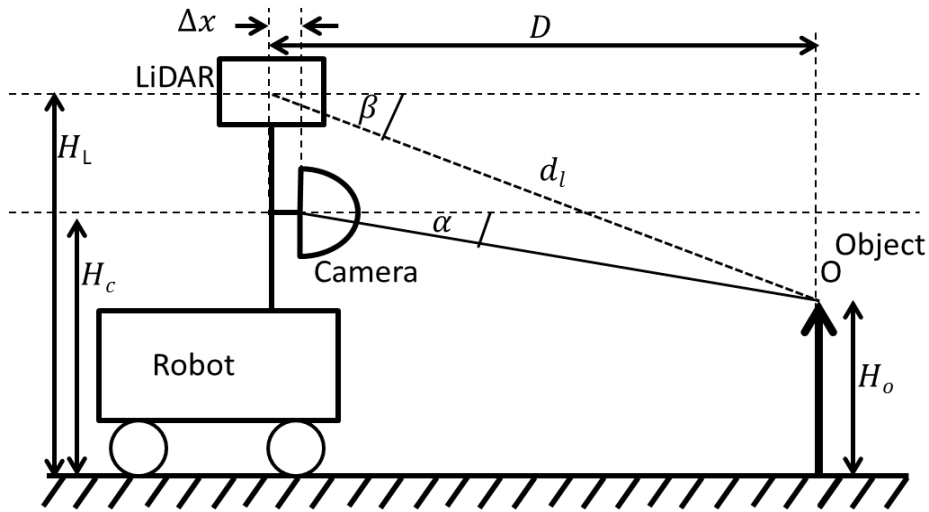

**Figure 2.** Side view of the sensor setup.

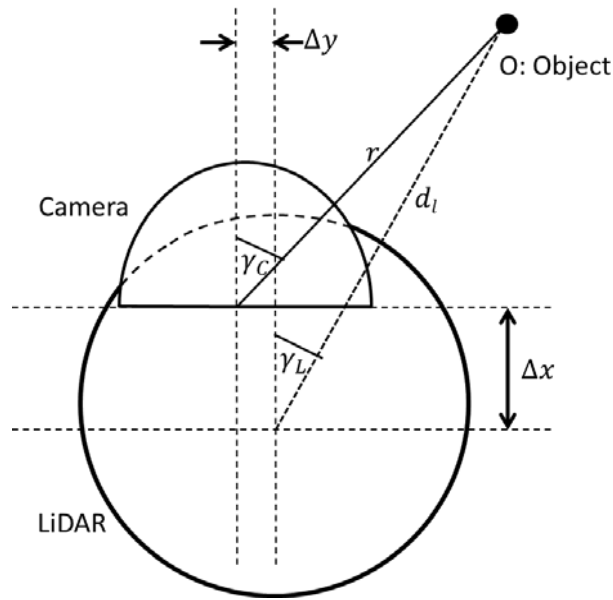

**Figure 3.** Top view of the sensor setup.

The notation used in Figures 2 and 3 is as follows:

$\Delta x$ = Frontal displacement of the centres of LiDAR and camera sensor. $\Delta y$ = horizontal displacement of the centres of LiDAR and camera sensor. $d_l$ = distance to the object O sensed by LiDAR. $\beta, \gamma_L$ = latitude and longitude of object O as measured by the LiDAR, respectively.

$H_C$ = height of the camera from the ground. $H_L$ = vertical height of the LiDAR, and $\alpha, \gamma_C$ = latitude and longitude of object O as measured by the camera, respectively.

The values $d_L, \beta$, and $\gamma_L$ are the outputs of the LiDAR sensor. The purpose of this alignment is to find the corresponding pixel in the camera output for each data point output by the LiDAR sensor. Here we assume that the main axis of the camera and the LiDAR are aligned with each other.

Considering the distance to object O, we have:

$$D = d_l \cos \beta \cdot \cos \gamma_L = r \cdot \cos \alpha \cdot \cos \gamma_C + \Delta x \tag{1}$$

Considering the vertical height of the object O, we have:

$$H_O = H_L - d_L \cdot \sin \beta = H_C - r \cdot \sin \alpha, \tag{2}$$

From (1) and (2), we can calculate corresponding latitude $\alpha$ of the camera as follows:

$$\tan \alpha = \frac{((H_C - H_L) + d_L \cdot \sin \beta) \cdot \cos \gamma_C}{d_l \cdot \cos \beta \cdot \cos \gamma_L - \Delta x} \tag{3}$$

Considering the horizontal displacement from the setup in Figure 3, we have:

$$d_l \cos \beta \cdot \sin \gamma_L = r \cdot \cos \alpha \cdot \sin \gamma_C + \Delta y \tag{4}$$

From (1) and (4), we can calculate corresponding longitude $\gamma_C$ of the camera as follows:

$$\tan \gamma_C = \frac{d_l \cdot \cos \beta \cdot \sin \gamma_L + \Delta y}{d_l \cdot \cos \beta \cdot \cos \gamma_L - \Delta x} \tag{5}$$

The Equations (3) and (5) pave the way to align the data points of the LiDAR and the camera. The purpose of the calibration process is to find the parameters $H_C, H_L, \Delta y, \Delta x$.

Although posing minimalistic needs for calibration, the above geometric alignment process cannot be fully relied upon as a robust mechanism, because errors in calibration measurements, imperfections in sensor assembly, and per-unit variations derived from the manufacturing processes may introduce factors that deviate from the ideal sensor geometry. For example, the curvature of the 360° camera might not be uniform across its surface. Therefore, to be robust enough for such discrepancies, the geometrically aligned data ideally must undergo another level of adjustment. This is accomplished in the next stage of the framework by utilizing the spatial correlations that exist in image data.

Another problem that arises when fusing data from different sources is the difference in data resolution. For the case addressed in this paper, the resolution of LiDAR output is considerably lower than the images from the camera. Therefore, the next stage of the data fusion algorithm is designed to match the resolutions of LiDAR data and imaging data through an adaptive scaling operation.

### 3.2. Resolution Matching Based on Gaussian Process Regression

In this section we describe the proposed mechanism to match the resolutions of LiDAR data and the imaging data. In Section 3.1, through geometric alignment, we matched the LiDAR data points with the corresponding pixels in the image. However, the image resolution is far greater than the LiDAR output. The objective of this step is to find an appropriate distance value for the image pixels for which there is no corresponding distance value. Furthermore, another requirement of this stage is to compensate for discrepancies or errors in the geometric alignment step.

We formulate this problem as a regression based missing value prediction, where the relationship between the measured data points (available distance values) is utilized to interpolate the missing values. For this purpose we use Gaussian Process Regression (GPR) [20], which is a non-linear regression technique. GPR allows to define the covariance of the data in any suitable way. In this step, we derive the covariance from the image data, and thereby adjusting to account for discrepancies in the geometric alignment stage. A Gaussian Process (GP) is defined as a Gaussian distribution over functions [20]:

$$f(x) \sim GP(m(x), \kappa(x, x')), \tag{6}$$

where, $m(x) = \mathbb{E}[f(x)]$, and $\kappa(x, x') = \mathbb{E}[(f(x) - m(x))(f(x) - m(x))^T]$.

The power of GPs lies in the fact that we can define any covariance function $\kappa$ as relevant to the problem at hand.

Let's denote a patch of size $n \times n$ extracted from the depth map D, as $y_i$. Pixels in an extracted patch is numbered from 1 to $n^2$—increasing along the rows and columns. Some pixels of this patch have a distance value associated with it (geometric alignment stage). The objective of this regression step is to fill the rest of the pixels with an appropriate depth value.

The pixels with a depth associated to it will act as the training set $D = \{(x_i, f_i), i = 1: N\}$, where N is the number of pixels that has a depth associated with it, and $x_i$ is the pixel number. Let $X = \{x_i, i = 1: N\}$, $f = \{f_i, i = 1: N\}$, and $X^* = \{x_j, j = 1: n^2 - N\}$, is the set of pixel numbers for which the depth map is empty. The resolution matching problem then becomes to find $f^* = \{f_j^*, j = 1: n^2 - N\}$, the depth of the pixels corresponding to $X^*$. By definition of the GP, the joint distribution between $f$ and $f^*$ has the following form:

$$\begin{pmatrix} f \\ f^* \end{pmatrix} \sim \mathcal{N}\left( \begin{pmatrix} \mu \\ \mu^* \end{pmatrix}, \quad \begin{pmatrix} K & K_* \\ K_*^T & K_{**} \end{pmatrix} \right) \tag{7}$$

where, $K = \kappa(X, X)$, $K_* = \kappa(X, X^*)$, and $K_{**} = \kappa(X^*, X^*)$, are the covariance matrices defined utilising the covariance function $\kappa$, $\mu$ and $\mu^*$ is the corresponding mean vectors for $f$ and $f^*$. The solution to $f^*$ is given as the posterior predictive density as follows [33]:

$$p(f^*|X^*, X, f) = \mathcal{N}(f^*|\mu^*, \Sigma^*) \tag{8}$$

where:

$$\mu^* = \mu(X^*) - K_*^T K_y^{-1}(f - \mu(X)) \tag{9}$$

and:

$$\Sigma^* = K_{**} - K_*^T K^{-1} K_* \tag{10}$$

A suitable covariance function $\kappa$ has to be defined to meet the objective of filling the missing values in the depth map D. So how do we define a suitable covariance function? To do so, we make the assumption that similar pixels of the colour image will have the same depth value. Similarity of the pixels is defined based on the Euclidian distance between the pixels and the grey-level of the pixel. As such we define the covariance function as, $\kappa(x, x') = c(x, x') \cdot s(x, x')$, where the covariance between any two pixels $x, x'$, $\kappa(x, x')$ is the multiplication of two factors: $c(x, x')$: closeness between the two pixels in terms of spatial Euclidian distance and $s(x, x')$: similarity between the two pixels in terms of its grey-level value, defined as follows:

$$c(x, x') = \exp\left( -\frac{1}{2} \cdot \frac{\|x - x'\|^2}{K_p} \right) \tag{11}$$

$$s(x, x') = \exp\left( -\frac{1}{2} \cdot \frac{(I_x - I_{x'})^2}{K_I} \right) \tag{12}$$

where $I_x$ denotes the grey-level value of the camera image at pixel position $x$. $K_p$ and $K_I$ controls the width of the respective kernels. The missing depth value of a pixel $x$, is taken to be the mean value $\mu^*$ at x given by Equation (9), and the corresponding uncertainty of the calculated pixel value is taken to be the variance at $x$ given by Equation (10). To summarize, the GP based regression to fill the missing depth values is illustrated in Figure 4.

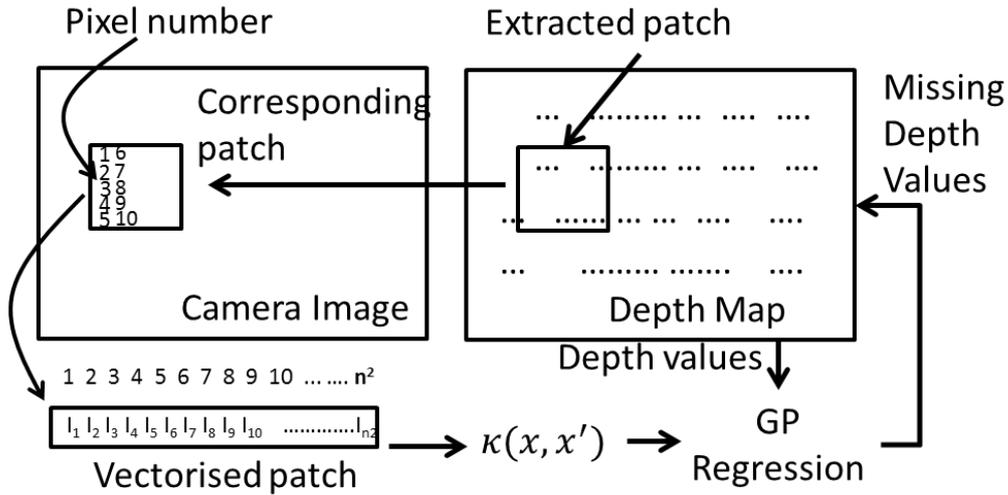

**Figure 4.** The sequence of steps in Gaussian Process (GP)-based resolution matching algorithm.

## 4. Experimental Results and Discussion

In this section we will describe the experimental test bed from which test data were gathered, and discuss the results obtained with the data fusion algorithm described in the previous section. This section is mainly organized in four parts. The first subsection will describe the test bed and the data set. The second subsection presents results for the GP based resolution matching algorithm and Subsection 4.3 will demonstrate the robustness of multimodal data fusion based Free Space Detection (FSD) algorithm as compared to FSD with individual sensors. Finally, Subsection 4.4 will discuss the overall results and limitations of the current work.

### 4.1. The Experimental Setup and Dataset

#### 4.1.1. Description of the Test Bed and Estimation of Extrinsic Parameters

To collect the data necessary for the experiment, a test bed was assembled as illustrated in Figure 5. The test bed is composed of a front facing wide angle camera, a rear facing camera, a radar, and a LiDAR scanner tagged to an electric quad bike. However, in this paper we are focused on fusing only the front facing wide angle camera output with the LiDAR scanner output.

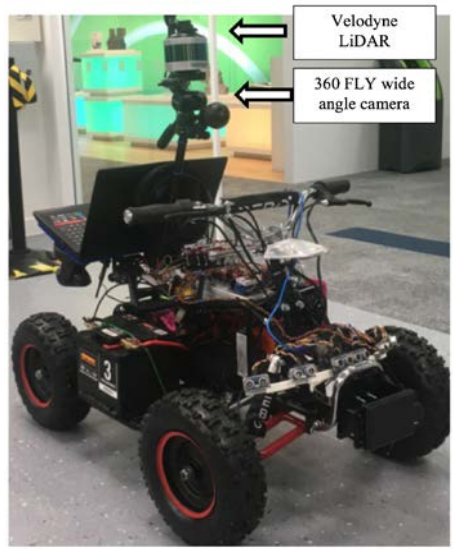

**Figure 5.** The experimental test bed used for data capture.

The VLP-16 LiDAR (Velodyne, San Jose, California, United States) which is used in the test bed, a compact low power light-weight optical sensor, has a maximum range of 100 m. The sensor supports 16 channel communications taking a total of 300,000 measurements per second. Data is captured over 360° on the horizontal axis and 30° on the vertical, utilizing 16 laser/detector pairs.

The wide-angle camera utilized in the setup is a 360Fly camera (360Fly Inc., Canonsburg, Pennsylvania United States) is enclosed in a 61 mm diameter sphere with a single fish eye lens mounted on the top. The field of view is 360°on the vertical and 240° horizontal. Standard 360° video that is output from this camera is a flat equirectangular video displayed as a sphere.

Extrinsic calibration of a monocular camera and a LiDAR sensor data is the process of estimating the disparity between the camera and the LiDAR. The parameters described in section 3 correspond to this disparity. The fundamental challenge of extrinsic calibration is when the camera-lidar sensors do not overlap or share the same field of view [34]. Motivated by the extrinsic calibration process presented in [35], in this research a single calibration target as shown in Figure 6a, with four circular holes was used. The centre of the circles lines up to form a single rectangle measuring 37.5 cm × 26.5 cm. The target was intended to be perceived by the sensors from a unique point of view, thus avoiding the need for multiple targets during the process. Each circle in Figure 6d acted as a distinct feature visible to both monocular camera and a LiDAR sensor. Both sensors field of view was overlapping, and the target was positioned so that a minimum of two LiDAR beams intersected with the circumference of the circle. After, an initial estimation of extrinsic parameters through measurement, parameters were adjusted until the circles on the LiDAR and the camera overlapped.

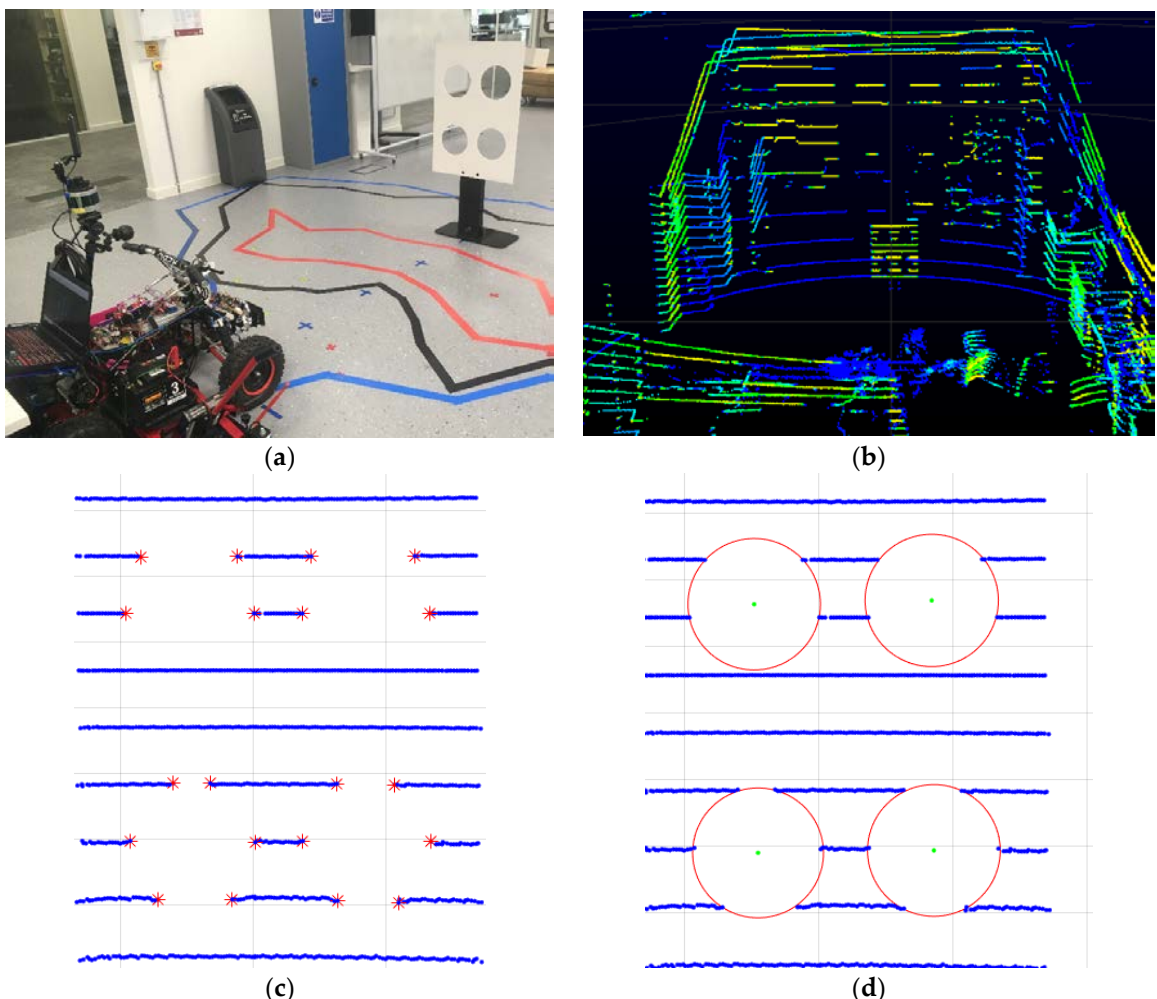

**Figure 6.** Steps in extrinsic calibration process. (**a**) Camera setup and calibration target; (**b**) The corresponding LiDAR capture; (**c**) Red points on the edges of the circle; (**d**) Identification of the circle centers.

Once the parameters are estimated using the calibration procedure, the LiDAR data points are projected on to the wide-angle image. An example projection is illustrated in Figure 7. The measurements for the geometric alignment stage are as follows: $H_C = 0.55$ m, $H_L = 0.61$ m, $\Delta y = 0.07$ m, $\Delta x = 0.5$ m.

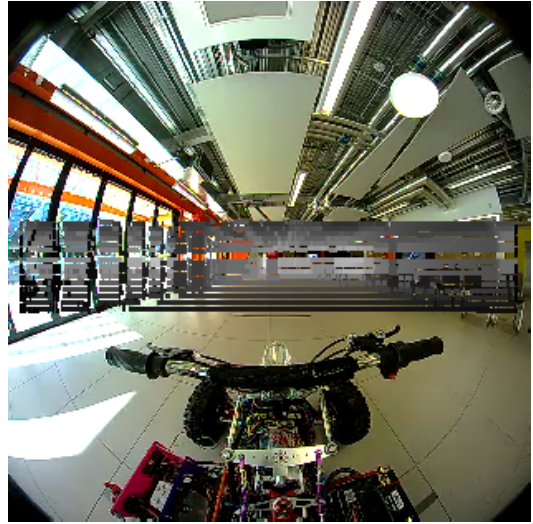

**Figure 7.** Projection of LiDAR data on to the spherical video frame.

### 4.1.2. Description of the Data Set

The experimental results are based on custom data captures utilizing the platform described in the previous subsection. Six datapoints covering a range of situations are captured to illustrate different conditions. Each datapoint consists of a data capture by the platform that moves in a straight line, until it cannot proceed due to an obstacle in its path. Each data capture is at least 10 s in length. The vehicle was moving at 5 km per hour (equivalent to 1.4 m/s), and the camera frame rate is 29 frames per second, while the LiDAR capture rate is 5 frames per second. The sensor data are stored and used in the subsequent processing. A screen capture of each datapoint used for experimental validation are illustrated in Figure 8.

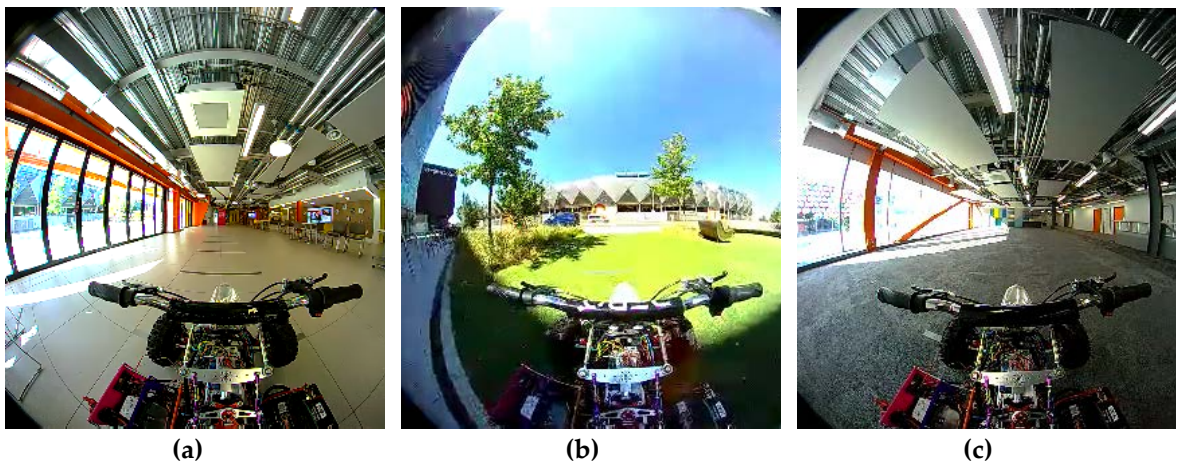

(a)          (b)          (c)

**Figure 8.** *Cont.*

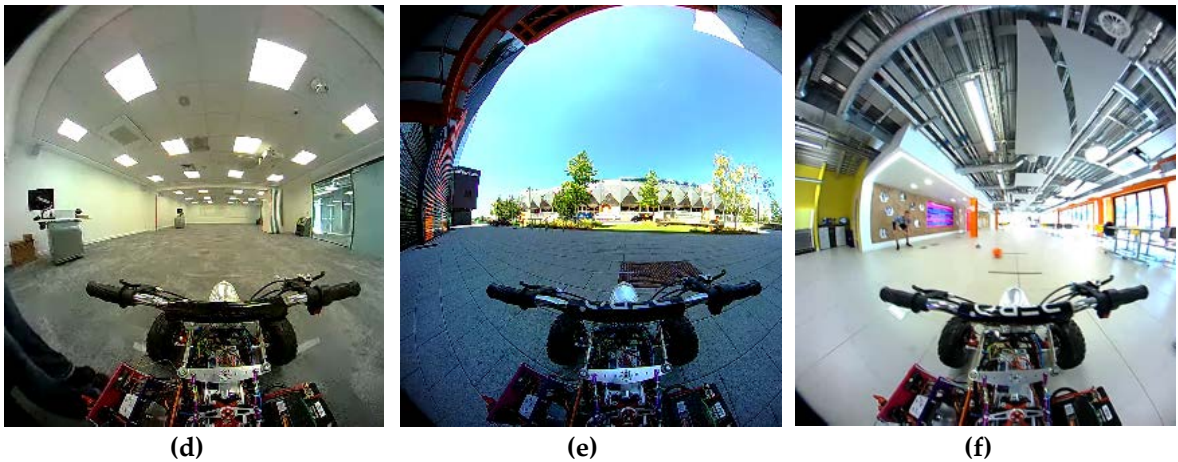

**Figure 8.** Screen shots of different scenarios captured by the quadbike in motion. (**a**) Scenario 1: Indoors tiled floor with cluttered environment; (**b**) Scenario 2: Outdoor environment on artificial grass; (**c**) Scenario 3: Indoor environment with carpet floor and extreme shadows; (**d**) Scenario 4: Indoor environment with less clutter; (**e**) Scenario 5: Outdoor environment with change of surface; (**f**) Scenario 6: Indoor environment with moving object in LiDAR blind spot.

### 4.1.3. Free Space Detection (FSD) as a Methodology for Evaluation of Robustness of Data Fusion

The purpose of data fusion is to assist subsequent data perception tasks. Hence, the performance of a data fusion framework need to be assessed within the context of subsequent processing stages. This subsection describes the methodology employed in this work to assess the performance of data fusion frameworks. To keep within the context of autonomous vehicles, we will utilize free space detection (FSD) as a representative perception task.

FSD is the mechanism by which an autonomous vehicle understands regions in the space to which it can move in to without bumping in to any obstacle. Typically, data driven learning methods are utilized to train FSD algorithms. In this experiment we illustrate the effectiveness of sensor data fusion in terms of training a FSD classifier.

As illustrated in Figure 9b, the data fusion stage produces a depth map indicating the distance from the LiDAR to each pixel in the colour image. In this study, free space is defined as any point in space, which is at the same level as the wheels of the test bed. We assume the surface on which the vehicle moves, is flat. Therefore, any pixel representing a point in space at the same level as of the bottom of the wheels, is considered a "free space" point.

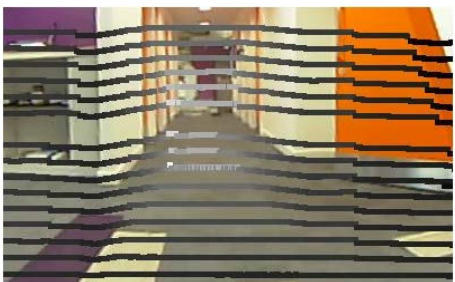 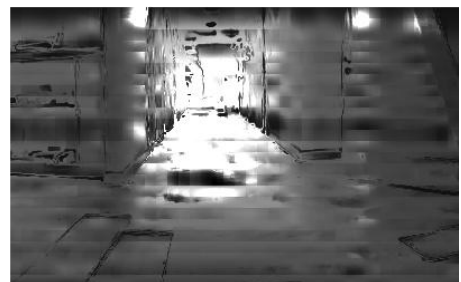

(**a**)                                           (**b**)

**Figure 9.** *Cont.*

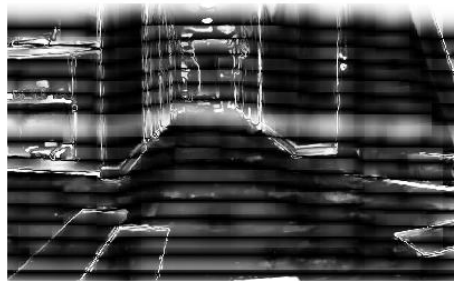

(**c**)

**Figure 9.** Visual illustration of the outputs of the Gaussian Process based resolution matching. (**a**) Aligned camera data and LiDAR points; (**b**) Output of Resolution matching step; (**c**) Uncertainty associated with depth estimation.

## 4.2. Evaluation of GP Based Resolution Matching Framework

### 4.2.1. Visual Illustration of Resolution Matching Results

Figure 9 provides a visual illustration of the inputs and outputs of the data fusion process. Figure 9a is the data samples to be fused. The markers on the image indicate the geometrically aligned distance data points. Figure 9b is the result of GP-based resolution matching step, and Figure 9c is the corresponding uncertainty associated with the depth value interpolations. Note that, to the middle of the images, there is a laser reading that is missing. Although GP-based regression fills those regions with reasonable values, the uncertainty is also high. Furthermore, some areas at which there are sharp colour discontinuities, there is a high level of uncertainty, for example on the carpet. Also note that data from one of the LiDAR scanners are missing in the middle of Figure 9a. In such situations, data fusion algorithm still finds a depth value, but it is represented with high level of uncertainty, as seen in Figure 9c.

### 4.2.2. Comparison of the Proposed GP Based Interpolation with other Competing Techniques

For performance evaluations that proceed, we utilize three image segments that are to be fused. Utilizing the fused output, which is a distance map as shown in Figure 9b, the free space points are identified. The logical mask that represents free space points is referred to as the "free space mask". The free space mask is then compared to the ground truth mask. The ground truth for the three image segments are manually marked. Similarly, to measure the performance of different techniques discussed in the proceeding sections, a free space detection mask is obtained and compared against the ground truth.

The number of pixels that do not match the ground truth is obtained by a simple 'xor' operation of the masks. The proportion of pixels that match the ground truth, which is referred to as "Accuracy" is used as the primary measure of performance. Furthermore, the precision and true positive rate is also calculated as measures of performance. These metrics are summarized in Table 1. We compare the performance of GP regression method with two other methods: tensor factorization for incomplete data [36] and robust smoothing based on discreet cosine transform [37]. The results are compared based on the performance of FSD algorithm, and are summarized in Table 1. As illustrated in Table 1, the proposed GP regression-based approach perform consistently well.

**Table 1.** Comparison of Different Resolution Matching Algorithms against the proposed Gaussian Process Framework.

| Algorithm | Accuracy | Precision | True positive |
|---|---|---|---|
| Proposed Gaussian Process Framework | 0.933 | 0.908 | 0.485 |
| Tensors factorization for missing values [36] | 0.8 | 0.827 | 0.383 |
| Robust DCT smoothing for grid data [37] | 0.923 | 0.921 | 0.467 |

Although recommended as a framework in [20], tensor factorization-based approaches such as [36] are not very suitable for high-dimensional imaging applications. The proposed tensor-based methods in literature are mainly targeted at low dimensional signals. Although slightly lower in performance compared to the proposed method in two test cases, the DCT based smoothing approach for grid data [37] performs well for the resolution matching scenario. This means the algorithm in [37] too is able to capture the high dimensional non-linear spatial variations. However, the downside of this approach is that it does not yield the uncertainty of the estimations.

### 4.3. Robust Free Space Detection Utilizing Data Fusion

The purpose of this section is to illustrate the usefulness of multimodal data fusion for robust image recognition tasks. In particular, we demonstrate the robustness of free space detection when utilizing data fusion. For this purpose, we will compare FSD with data fusion to FSD with individual sensors. In the next subsection we will briefly describe the two competing algorithms.

#### 4.3.1. Image Based FSD and LiDAR Based FSD.

To consider the performance improvements gained by sensor data fusion as compared to single sensor, we utilize two algorithms: one based only on camera data (image-based FSD) and one based only on LiDAR data (LiDAR-based FSD).

The image-based FSD algorithm is a supervised FSD algorithm that learns FS from examples. For this purpose, we collect a training set of image patches from the camera image and assign appropriate labels as free space or not. 1200 image patches of size $16 \times 16$ are collected from the 8 example video frames to train a Support Vector Machine (SVM) classifier. Histogram of Oriented Gradients (HoG) features are extracted from these training image patches, and the SVM is trained from those feature vectors. The HOG features are calculated for every $8 \times 8$ block within the $16 \times 16$ patch. The Radial Basis Function is used as the kernel for the SVM. 10-fold cross validation is utilized for model selection. The inbuilt functions of MATLAB®R2018a are used for feature extraction and training of the SVM ('fitcsvm' with default parameters). The LiDAR-based FSD algorithm utilized for this study is known as the Occupancy Grid Maps (OGMaps) [38]. In OGMaps, all the points in the 3-D point cloud of LiDAR scan is mapped on to a 2D grid.

#### 4.3.2. Comparison of Different Algorithms for FSD

The purpose of this experiment is to compare between different FSD algorithms. The image-based FSD and LiDAR-based FSD are illustrated and compared against the proposed FSD based on image and LiDAR fusion. Firstly, the occupancy grid maps based on LiDAR, image classification and for the fused approach are illustrated for several scenarios in Figure 10. The white rays emanating from the robot position, correspond to grid points that are free to roam. Here, for comparison purposes an OGMap is calculated for image-based FSD as follows: Firstly, the block-based image classifier is applied on the image frame. Secondly, the pixels that are considered not-free space are translated on to the X-Y plane. For this step, we assume the non-free space pixels correspond to an obstacle placed at the same ground level of the robot. The OGMap for the fused approach can be obtained by overlaying the LiDAR OGMap and image-based OGMap, and in the resultant OGMap the closest obstacle to the robot is considered during free space calculation. However, in this method, the uncertainty of the data sources are not captured, and the OGMap fusion takes a conservative approach. The proposed OGMap fusion goes one step beyond to consider the uncertainty of the data sources during fusion. This is done by considering the image based FSD results as confident only in the LiDAR blind spots. Similarly ascribing higher confidence to LiDAR OGMap only in the vertical space that it's laser beams cover allows to fuse the OGMaps in a better way. The fusion results of our proposed method is illustrated in Figure 11.

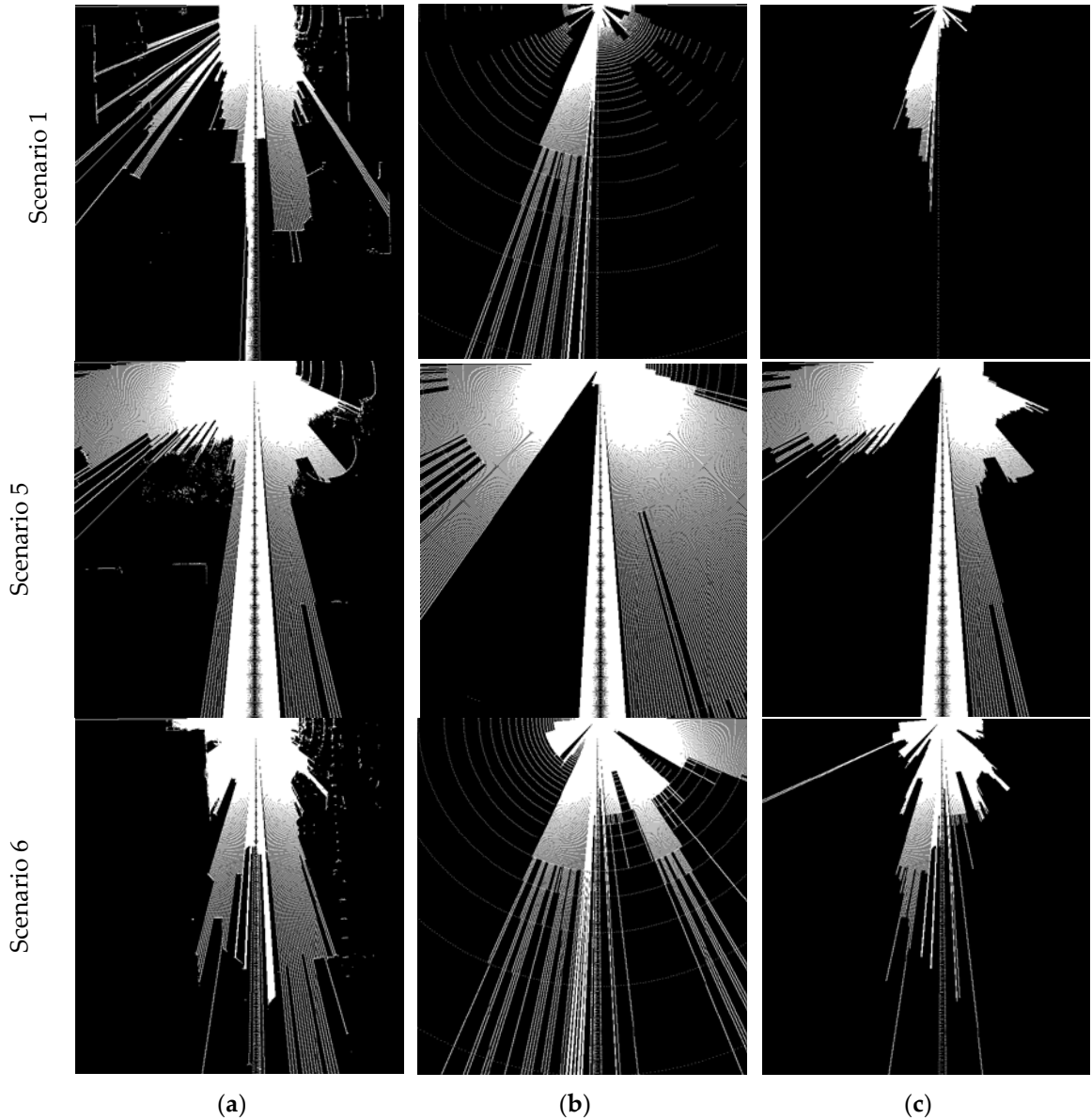

**Figure 10.** Comparison of free-space maps for different algorithms for three different scenarios. (**a**) LiDAR only; (**b**) Image classifier only; (**c**) Fusion-based approach.

The free space illustrated ass OGMaps as given in Figure 10 can be translated on to the corresponding image frames in Figure 11. The benefits of the proposed approach for FSD based on fusion can be illustrated as in Figure 11. For example, in Figure 11a, for Scenario 1, when the image-based classifier fails to detect extremely bright area as free space, the fusion approach will also consider it as non-free space. This area corresponds to a space in the LiDAR's blind spot. Similarly, for Scenario 6 in Figure 11, the ball that is thrown across the vehicle falls on the LiDAR's blind spot. However, the ball is captured as an obstacle by the image-based FSD algorithm, and when the results are fused, the ball is captured as an obstacle. Scenario 1 corresponds to a situation where the image based FSD is not performing very well. In this example, due to high saturation and mirroring on the floor, the area just next to the glass windows is not classified as free space (i.e., the classifier fails). However, the LiDAR information in this region shows high confidence and hence the fused image combines the two areas based on uncertainty. Scenario 5 illustrates a different situation where, due to an air-borne obstacle (eg., the experimenter's hand) that is visible only by the LiDAR, the FSD by LIDAR projects to see an obstacle on the floor. However, the image-based FSD performs very well on this occasion and the fused image demonstrates the best of the individual scenarios.

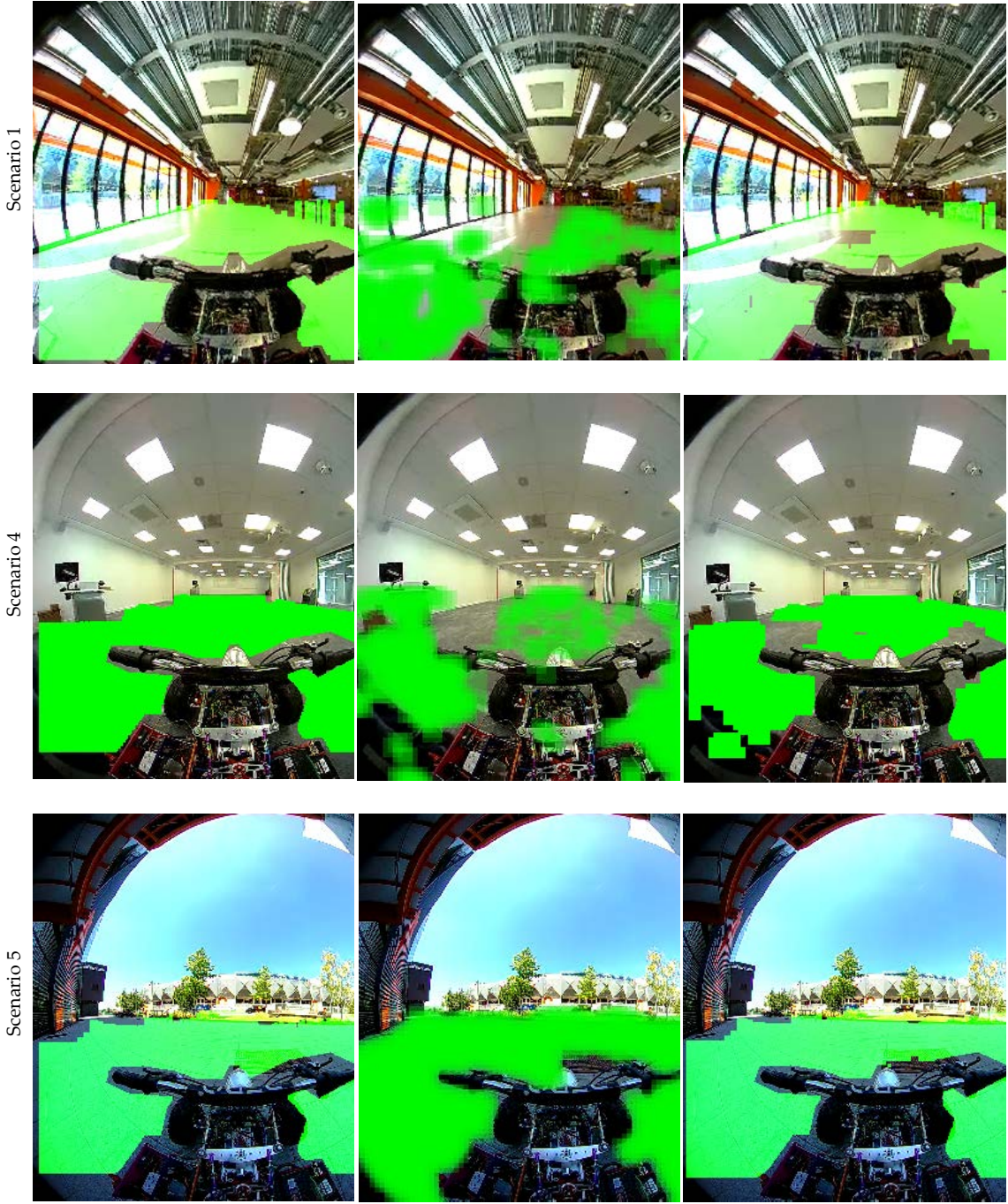

**Figure 11.** *Cont.*

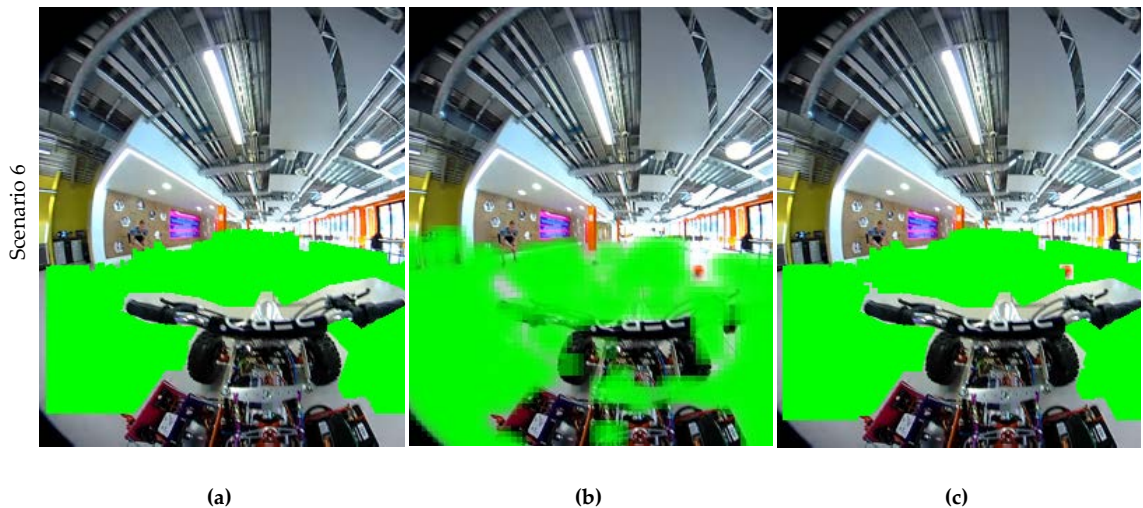

**(a)**             **(b)**             **(c)**

**Figure 11.** Comparison of free-space maps for different algorithms for three different scenarios. (**a**) LiDAR only; (**b**) Image classifier only; (**c**) Fusion based approach.

### 4.4. Discussion and Limitations

Multimodal data provides the opportunity to utilize the diversity offered by heterogenous physical sensors to overcome the limitations of individual sensors. The above sections demonstrated that data fusion will lead to more robust recognition algorithms. The challenges of data fusion go beyond extrinsic calibration. In the above sections we discussed how to overcome resolution mismatches in heterogenous data sources. Furthermore, different data sources have different uncertainties associated with them. We demonstrated how uncertainties associated with different data sources can be accounted for, to make better use of fused data for recognition tasks.

#### 4.4.1. Uncertainties Associated with Motion and Vibration of the Platform

While uncertainties in data can be caused by many factors, such as sensor malfunctions and imperfections, the most prevalent form of uncertainty in our system is due to artefacts caused by movement of the platform. In the following illustration, we shall demonstrate two scenarios with a static surrounding while the vehicle is in motion, and the LiDAR frames within the past 1 s time window are combined temporally to generate the occupancy grid map. The occupancy grid maps for Scenario 1 and Scenario 5 are illustrated in Figure 12. Scenario 5 corresponds to an outdoor paved environment with relatively higher vibration compared to indoor scenarios such as Scenario 1. However, considering the spread of depth values for static objects do not show significant difference between the two scenarios.

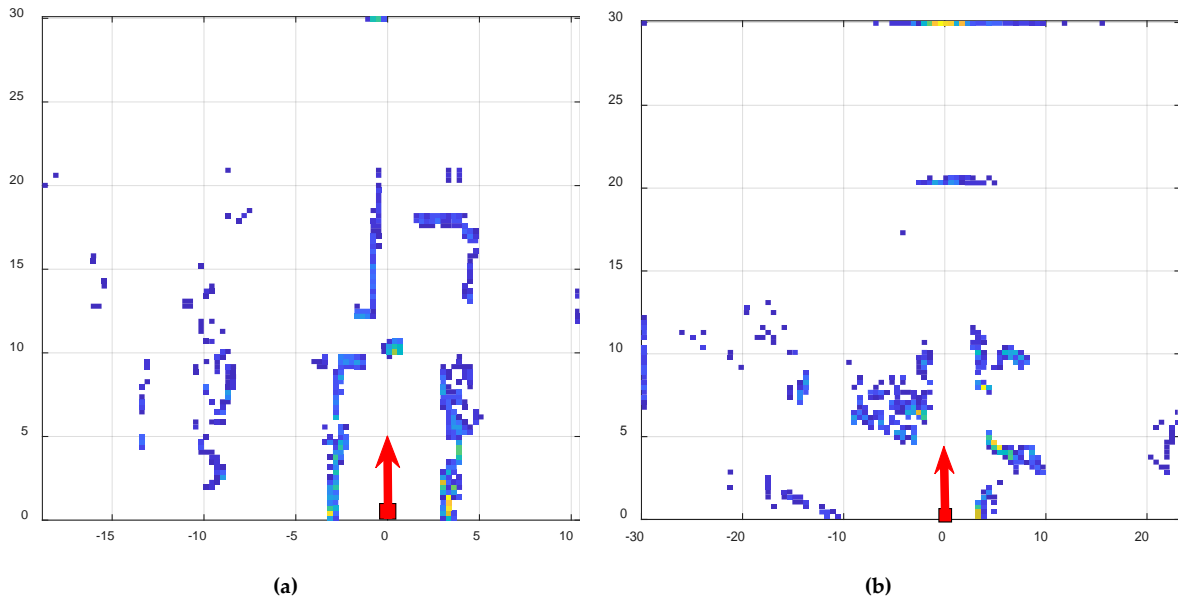

**Figure 12.** Occupancy Grid Map based on temporally combined LiDAR data. (a) Scenario 1; (b) Scenario 5.

The errors tend to accumulate when we consider a larger time window. For example, instead of 1 s, when the LiDAR frames in past 2 s are combined, the spread of values have increased from 0.75 m to 1.5 m. This is illustrated in Figure 13 for a subset of Scenario 1. Furthermore, while the vehicle is in motion, the video quality becomes poor due to motion artefacts. The quality of images captured are degraded due to the motion of the platform. The difference in a similar frame captured when the vehicle is in motion and not in motion is illustrated in Figure 14. The implication of this is that when the image based free space algorithm is trained, it should be trained on both the still image frames as well as on image frames when the vehicle is in motion.

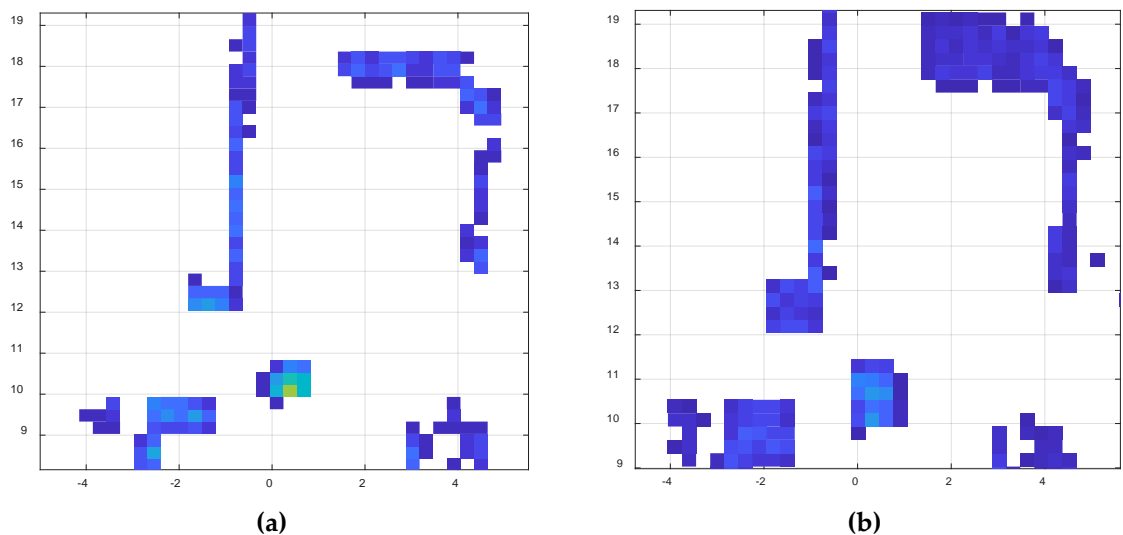

**Figure 13.** The effect of temporal window width for LiDAR frame accumulation. As the temporal window width is changed from 1s to 2s, the spread of values increases. **(a)** when temporal window width is 1s; **(b)** when temporal window width is 2s.

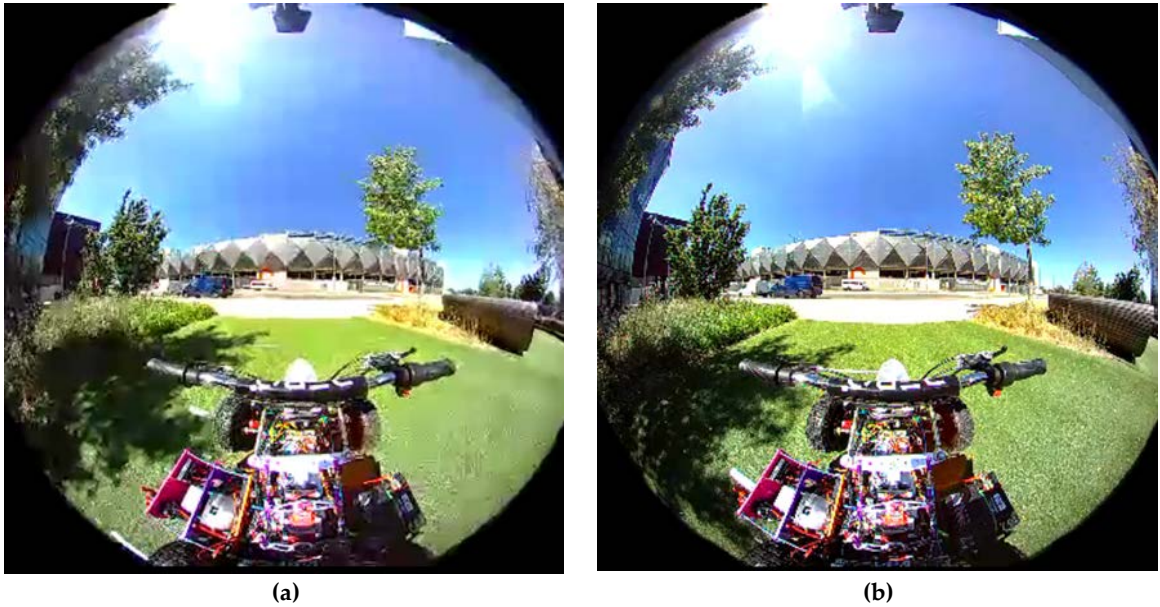

**(a)**                                                      **(b)**

**Figure 14.** The effect of motion artefacts on video capture. (**a**) When the vehicle in motion; (**b**) When the vehicle is still.

### 4.4.2. Limitations and Directions of Future Work

We identify several limitations of the proposed method for fusion and experimental setup. Firstly, the image-based FSD algorithm presented in this paper is utilized for demonstration of the benefits of fusion of LiDAR and camera data. It should be noted that, in terms of performance there are many advanced algorithms that are based on deep neural networks [39,40] or Gaussian Mixture Models [41]. The classifier used in this study is trained on a very small amount of data, as compared to state-of-the-art algorithms that are trained on thousands of hours of training data. However, image based free space detection and obstacle avoidance algorithms have well known failures such as when image pixels are saturated. The limitations of image based recognition algorithms are partly culpable for recent fatal accidents involving driverless vehicles [42,43].

Furthermore, the proposed data fusion method does not adequately address the issue of seeing through glass. Both types of the optical sensors used in this study, would not recognize see through glass as an object but identifies objects beyond the see-through glasses. This would become problematic for applications such as FSD. The reflectivity data that are generated in the LiDAR, which is not utilized in this method, might prove beneficial towards solving this issue. Another possible approach is to utilize other sensors such as ultra sound sensors.

The advantage of the GP-based interpolation is that it calculates the uncertainty of the interpolated values. The Gaussian process interpolation algorithm presented in this paper relies on a covariance function that takes in to account both the Euclidean distance and the pixel-luminance similarity. Therefore, the algorithm assumes that similar coloured pixels around a known pixel (a pixel for which we know the distance) will have a similar distance from the LiDAR. This is one of the main limitations of the current work. For example, the floor tiles will have a similar luminance value, and hence the distance output from the GP based interpolation algorithm will yield almost the same distance. However, the actual distance gradually changes along the floor, and this attribute can only be incorporated in to the GP framework by utilizing a covariance function that captures the surface gradient of the objects. We will be investigating such a covariance function in our future work.

The current experimental setup uses a LiDAR scanner with 16 lasers. For this reason, the vertical resolution of the LiDAR scanner is quite limited. It should be noted that denser LiDAR scanners are now commercially available, such as VLP-64, which has 64 lasers and VLS-128 which has 128 lasers. Such LiDAR scanners make it possible for dense depth maps to be captured and processed. However, as of the time of writing, such dense laser scanners are not mass produced, and are extremely expensive. Therefore, especially for low cost applications, at least in the shorter term, the low density laser scanners will have to be utilized.

## 5. Conclusions and Future Work

This paper addresses the problem of fusing the outputs of a LiDAR scanner and a wide-angle monocular image sensor. The first part of the proposed framework spatially aligns the two sensor data streams with a geometric model. The resolutions of the two sensors are quite different, with the image sensor having a much denser spatial resolution. The two resolutions matched, in the second stage of the proposed framework, by utilizing a Gaussian Process regression algorithm that derives the spatial covariance from the image sensor data. The output of the GP regression not only provides an estimation of the corresponding distance value of all the pixels in the image, but also indicates the uncertainty of the estimation by way of standard deviation. The advantages of the proposed data fusion framework is illustrated through performance analysis of a free space detection algorithm. It was demonstrated that perception tasks in autonomous vehicles/mobile robots can be significantly improved by multimodal data fusion approaches, as compared to single sensor-based perception capability. As compared to extrinsic calibration methods, the main novelty of the proposed approach is the ability to fuse multimodal sensor data by accounting for different forms of uncertainty associated with different sensor data streams (resolution mismatches, missing data, and blind spots). The future work planned, includes extension of the sensor fusion framework to include multiple cameras, radar scanners and ultra sound scanners. Furthermore, we will research methods for robust free space detection based on the data fusion framework. Advanced forms of uncertainty quantification techniques will be utilized to capitalize on the diversity offered by multimodal sensors.

**Author Contributions:** For research articles with several authors, a short paragraph specifying their individual contributions must be provided. The following statements should be used "Conceptualization, Varuna De Silva and Ahmet Kondoz; Methodology, Varuna De Silva, Jamie Roche; Software, Varuna De Silva; Validation, Varuna De Silva, Jamie Roche.; Resources, Ahmet Kondoz, Jamie Roche.; Data Curation, Varuna De Silva; Writing-Original Draft Preparation, Varuna De Silva.; Writing-Review & Editing, Varuna De Silva, Jamie Roche.; Funding Acquisition, Ahmet Kondoz, Varuna De Silva.", please turn to the CRediT taxonomy for the term explanation. Authorship must be limited to those who have contributed substantially to the work reported.

**Acknowledgments:** Authors would like to thank Loughborough University for providing seed funding for the research carried out and for the doctoral studentship of Jamie Roche.

## References

1. Park, S.; Choi, W.; Han, D.K.; Ko, H. Acoustic event filterbank for enabling robust event recognition by cleaning robot. *IEEE Trans. Consum. Electron.* **2015**, *61*, 189–196.
2. Lumpkins, W. Driverless Cars and Driverless Vacuums: Will the Madness Never End? [Product Reviews]. *IEEE Consum. Electron. Mag.* **2014**, *3*, 88–91.
3. Zhang, J.; Song, G.; Qiao, G.; Meng, T.; Sun, H. An indoor security system with a jumping robot as the surveillance terminal. *IEEE Trans. Consum. Electron.* 2011. 57, 1774–1781.
4. Mechsy, L.S.R.; Dias, M.U.B.; Pragithmukar, W.; Kulasekera, A.L. A mobile robot based watering system for smart lawn maintenance. In Proceedings of the 2017 17th International Conference on Control, Automation and Systems (ICCAS), Jeju, Korea, 18–21 October 2017; pp. 1537–1542.
5. Sahin, H.; Guvenc, L. Household robotics: autonomous devices for vacuuming and lawn mowing [Applications of control]. *IEEE Control Syst.* **2007**, *27*, 20–96.
6. Webster, M., Dixon, C., Fisher, M., Salem, M., Saunders, J., Lee Koay, K., Dautenhahn, K., Saez-Pons, J. Toward Reliable Autonomous Robotic Assistants Through Formal Verification: A Case Study. *IEEE Trans. Hum.-Mach. Syst.* **2016**, *46*, 186–196.
7. Markwalter, B. The Path to Driverless Cars [CTA Insights]. *IEEE Consum. Electron. Mag.* **2017**, *6*, 125–126.

8. Jones, L. Driverless when and cars: Where? *Eng. Technol.* **2017**, *12*, 36–40.

9. Broggi, A.; Cerri, P.; Debattisti, S., Chiara Laghi, M., Medici, P., Panciroli, M., Prioletti, A. PROUD-Public Road Urban Driverless-Car Test. *IEEE Trans. Intell. Transp. Syst.* **2015**, *16*, 3508–3519.

10. Rasshofer, R.H.; Spies, M.; Spies, H. Influences of weather phenomena on automotive laser radar systems. *Adv. Radio Sci.* **2011**, *9*, 49–60.

11. Kytö, M.; Nuutinen, M.; Oittinen, P. Method for measuring stereo camera depth accuracy based on stereoscopic vision. In Proceedings of the 2011 Three-Dimensional Imaging, Interaction, and Measurement, San Francisco Airport, CA, USA, 24–27 January 2011.

12. Luo, R.C.; Yih, C.-C.; Su, K.L. Multisensor fusion and integration: approaches, applications, and future research directions. *IEEE Sens. J.* **2002**, *2*, 107–119.

13. Choi, B.S.; Lee, J.J. Sensor network based localization algorithm using fusion sensor-agent for indoor service robot. *IEEE Trans. Consum. Electron.* **2010**, *56*, 1457–1465.

14. Dan, B.K.; Kim, Y.S.; Suryanto; Jung, J.Y.; Ko, S.J. Robust people counting system based on sensor fusion. *IEEE Trans. Consum. Electron.* **2012**, *58*, 1013–1021.

15. Bai, J.; Lian, S.; Liu, Z.; Wang, K.; Liu, D. Smart guiding glasses for visually impaired people in indoor environment. *IEEE Trans. Consum. Electron.* **2017**, *63*, 258–266.

16. Erden, F.; Çetin, A.E. Hand gesture based remote control system using infrared sensors and a camera. *IEEE Trans. Consum. Electron.* **2014**, *60*, 675–680.

17. Zhang, J.; Singh, S. Visual-lidar odometry and mapping: Low-drift, robust, and fast. In Proceedings of the 2015 IEEE International Conference on Robotics and Automation (ICRA), Seattle, WA, USA, 26–30 May 2015, pp. 2174–2181.

18. Li, J. Fusion of Lidar 3D Points Cloud with 2D Digital Camera Image. Oakland University: Rochester, MI, USA, 2015.

19. Li, J.; He, X.; Li, J. 2D LiDAR and camera fusion in 3D modeling of indoor environment. In Proceedings of the 2015 National Aerospace and Electronics Conference (NAECON), 15–19 June 2015, Dayton, OH, USA, pp. 379–383.

20. Lahat, D.; Adali, T.; Jutten, C. Multimodal data fusion: an overview of methods, challenges, and prospects. *Proc. IEEE* **2015**, *103*, 1449–1477.

21. Gilani, S.A.N.; Awrangjeb, M.; Lu, G. Fusion of LiDAR data and multispectral imagery for effective building detection based on graph and connected component analysis. *Int. Arch. Photogramm. Remote Sens. Spatial Inf. Sci.* **2015**, *XL-3/W2*, 65–72.

22. Gong, X.; Lin, Y.; Liu, J. Extrinsic calibration of a 3D LiDAR and a camera using a trihedron. *Opt. Lasers Eng.* **2013**, *51*, 394–401.

23. Gong, X.; Lin, Y.; Liu, J. 3D LiDAR-Camera Extrinsic Calibration Using an Arbitrary Tribedron. *Sensors* **2013**, *13*, 1902–1918.

24. Alismail, H.; Baker, D.L.; Browning, B. Automatic calibration of a range sensor and camera system. In Proceedings of the 3DiMPVT, Seattle, WA, USA, 29 June–1 July 2013.

25. Park, Y. Calibration between color camera and 3D LiDAR instruments with a polygonal planar board. *Sensors* **2014**, *14*, 3, 5333–5353.

26. Moreno, G.; Ivan, A. LiDAR and panoramic camera extrinsic calibration approach using a pattern plane. *Pattern Recognit.* **2013**, *7914*, 104–113.

27. Lipu, Z. A new minimal solution for the extrinsic calibration of a 2D LiDAR and a camera using three plane-line correspondences. *IEEE Sens. J.* **2014**, *14*, 442–454.

28. Lipu, Z.; Deng, Z. Extrinsic calibration of a camera and a LiDAR based on decoupling the rotation from the translation. In Proceedings of the Intelligent Vehicles Symposium (IV), Madrid, Spain, 3–7 June 2012.

29. Lipu, Z.; Deng, Z. A new algorithm for the extrinsic calibration of a 2D LiDAR and a camera. *Meas. Sci. Technol.* **2014**, *25*, 065107.

30. Mastin, A.; Kepner, J.; Fisher, J. Automatic registration of LiDAR and optical images of urban scenes. In Proceedings of the Computer Vision and Pattern Recognition, Miami, FL, USA, 22–24 June 2009; pp. 2639–2646.

31. Maddern, W.; Newman, P. Real-time probabilistic fusion of sparse 3D LiDAR and dense stereo. In Proceedings of the 2016 IEEE/RSJ International Conference on Intelligent Robots and Systems (IROS), Daejeon, Korea, 9–14 October 2016; pp. 2181–2188.

32. Abbas, S.M.; Muhammad, A. Outdoor RGB-D SLAM performance in slow mine detection. In Proceedings of ROBOTIK, Munich, Germany, 21–22 May.

33. Murphy, K.P. *Machine Learning : A Probabilistic Perspective*. MIT Press: Cambridge, MA, USA, 2012.

34. Pandey, G.; McBride, J.R.; Savarese, S.; Eustice, R.M. Automatic extrinsic calibration of vision and LiDAR by maximizing mutual information. *J. Field Rob.* **2015**, *32*, 696–722. 2015.

35. Guindel, C.; Beltrán, J.; Martín, D.; García, F. Automatic Extrinsic Calibration for Lidar-Stereo Vehicle Sensor Setups. In Proceedings of the 2017 IEEE 20th International Conference on Intelligent Transportation Systems (ITSC), Yokohama, Japan, 16–19 October 2017.

36. Acar, E.; Dunlavy, D.M.; Kolda, T.G.; Mørup, M. Scalable tensor factorizations for incomplete data. *Chemom. Intell. Lab. Syst.* **2011**, *106*, 41–56.

37. Garcia, D. Robust smoothing of gridded data in one and higher dimensions with missing values. *Comput. Stat. Data Anal.* **2010**, *54*, 1167–1178.

38. Probabilistic Robotics. Available online: https://mitpress.mit.edu/books/probabilistic-robotics. (accessed on 19 July 2018).

39. Sanberg, W.P.; Dubbelman, G.; de With, P.H.N. Free-Space Detection with Self-Supervised and Online Trained Fully Convolutional Networks. *Electron. Imaging.* **2017**, *2017*, 54–61.

40. Levi, D.; Garnett, N.; Fetaya, E. StixelNet: A Deep Convolutional Network for Obstacle Detection and Road Segmentation. In Proceedings of the 2015 British Machine Vision Conference, Swansea, UK, 7–10 September 2015.

41. Yao, J.; Ramalingam, S.; Taguchi, Y.; Miki, Y.; Urtasun, R. Estimating drivable collision-free space from monocular video. In Proceedings of the 2015 Applications of Computer Vision (WACV), Waikoloa Beach, HI, USA, 6–9 January 2015; pp. 420–427.

42. Stewart, J. Tesla's Self-Driving Autopilot Involved in Another Deadly Crash. Available online: Link https://www.wired.com/story/tesla-autopilot-self-driving-crash-california/ (accessed on 18/ 07/ 2018).

43. Stewart, J. Why Tesla's Autopilot Can't See a Stopped Firetruck. Available online: Link https://www.wired.com/story/tesla-autopilot-why-crash-radar/ (accessed on 18/07/2018)